\documentclass{article}
\usepackage{arxiv_icml-like}

\usepackage{amssymb}
\usepackage{amsmath}
\usepackage{float}
\usepackage{graphicx}
\graphicspath{{./Images/}}
\usepackage{bm}
\usepackage[ruled,vlined]{algorithm2e}
\usepackage{multirow}
\usepackage{multicol}
\usepackage{siunitx}
\usepackage{booktabs}
\usepackage{mathtools}
\usepackage{xurl}
\usepackage{cuted}
\usepackage{stfloats} 
\newcommand{\mat}[1]{\bm{#1}}
\renewcommand{\vec}[1]{\mathbf{#1}}
\newcommand{\ten}[1]{\text{\sffamily\bfseries\slshape #1\/}}

\usepackage{listings}
\definecolor{codegreen}{rgb}{0,0.4,0}
\definecolor{codegray}{rgb}{0.4,0.4,0.4}
\definecolor{codepurple}{rgb}{0.5,0,0.7}
\definecolor{backcolour}{rgb}{0.96,0.96,0.94}
 
\lstdefinestyle{mystyle}{
    backgroundcolor=\color{backcolour},   
    commentstyle=\color{codegreen},
    keywordstyle=\color{magenta},
    numberstyle=\tiny\color{codegray},
    stringstyle=\color{codepurple},
    basicstyle=\ttfamily\footnotesize,
    breakatwhitespace=false,         
    breaklines=true,                 
    captionpos=b,                    
    keepspaces=true,                 
    numbers=left,                    
    numbersep=5pt,                  
    showspaces=false,                
    showstringspaces=false,
    showtabs=false,                  
    tabsize=2
}
 
\titlerunning{Graph Network-based Structural Simulator}
\begin{document}

\twocolumn[

\title{Graph Network-based Structural Simulator: Graph Neural Networks for Structural Dynamics}

\icmlsetsymbol{corr}{*}

\begin{authorlist}
\author{Alessandro Lucchetti}{polimi}
\author{Francesco Cadini}{polimi}
\author{Marco Giglio}{polimi}
\author{Luca Lomazzi}{polimi,corr}

\end{authorlist}

\affiliation{polimi}{Politecnico di Milano, Department of Mechanical Engineering, Via La Masa 1, Milano, 20156, Italy}

\correspondingauthor{}{{luca.lomazzi@polimi.it}}
\keywords{Graph Neural Network; Structural dynamics; Wave; Structural health monitoring; Surrogate Model}
]

\printAffiliations{} %

\begin{strip}
\begin{abstract}
Graph Neural Networks (GNNs) have recently been explored as surrogate models for numerical simulations. While their applications in computational fluid dynamics have been investigated, little attention has been given to structural problems, especially for dynamic cases. To address this gap, we introduce the \textit{Graph Network-based Structural Simulator (GNSS)}, a GNN framework for surrogate modeling of dynamic structural problems.  

GNSS follows the encode--process--decode paradigm typical of GNN-based machine learning models, and its design makes it particularly suited for dynamic simulations thanks to three key features: (i) expressing node kinematics in node-fixed local frames, which avoids catastrophic cancellation in finite-difference velocities; (ii) employing a sign-aware regression loss, which reduces phase errors in long rollouts; and (iii) using a wavelength-informed connectivity radius, which optimizes graph construction. 

We evaluate GNSS on a case study involving a beam excited by a $50$\,kHz Hanning-modulated pulse. The results show that GNSS accurately reproduces the physics of the problem over hundreds of timesteps and generalizes to unseen loading conditions, where existing GNNs fail to converge or deliver meaningful predictions.  

Compared with explicit finite element baselines, GNSS achieves substantial inference speedups while preserving spatial and temporal fidelity. These findings demonstrate that locality-preserving GNNs with physics-consistent update rules are a competitive alternative for dynamic, wave-dominated structural simulations.
\end{abstract}
\end{strip}

\section{Introduction}
\label{sec:introduction}
High-fidelity simulations underpin design and decision-making across many areas of computational science and engineering. However, resolving transient dynamics with fine spatial meshes is still expensive when long horizons, high frequencies, or repeated solves over parameter sets are required. Surrogate models have therefore become popular as a means to accelerate inference while retaining accuracy where it matters, from fluids and granular flow to materials and mechanics~\citep{forrester2008engineering,herrmann2024deeplearning}.

Several approaches to surrogate modeling exist: classical Reduced-Order Models (ROMs) compress dynamics via low-rank bases (e.g., POD/Galerkin) or regression in latent coordinates~\citep{benner2015survey,quarteroni2016reduced}; Gaussian-process and Kriging surrogates excel with limited data and quantified uncertainty~\citep{rasmussen2006gaussianprocesses,sacks1989design,forrester2008engineering}; Deep Learning (DL) surrogates, ranging from convolutional encoder-decoder models to sequence models and neural operators, aim to learn solution operators or time-advancement rules directly from data, offering mesh- and geometry-agnostic generalization in favorable settings, and they scale to large datasets and complex, nonlinear regimes~\citep{ronneberger2015unet,shi2015convlstm,li2020fourier,lu2021deeponet,kovachki2023neural}. 

However, each class of approaches has its own inherent limitations. ROMs can struggle with strongly nonlinear phenomena and often require intrusive projection or stabilization~\citep{benner2015survey,quarteroni2016reduced}. Neural operators and Physics-Informed Neural Networks (PINNs) have shown strong results for wave propagation and inverse problems, yet they frequently rely on global transforms or explicit PDE supervision~\citep{li2020fourier,raissi2019physicsinformed,cuomo2022pinnreview}. At high frequency, global spectral layers must resolve short wavelengths, and PINN training can become stiff due to competing loss terms and the need for accurate residuals on fine collocation sets~\citep{kovachki2023neural,krishnapriyan2021characterizing,wang2022respecting,maddu2021inverse}. Convolutional Neural Networks (CNNs) are particularly suitable for grid-like data, but become cumbersome when simulating generic domains with complex shapes~\citep{bronstein2017geometric}.

Recently, a promising alternative to the models discussed above has emerged in the form of Graph Neural Networks (GNNs). They provide a powerful framework for surrogate modeling on both structured and unstructured meshes and possess several properties that make them particularly suitable for representing physical phenomena. For instance, they naturally incorporate locality by aggregating information from neighboring nodes and support weight sharing in a way that is inherently invariant to node ordering~\citep{gilmer2017neuralmessagepassing,zaheer2018deepsets,battaglia2018relational}. These features align closely with the behavior of most physical systems, where locality and invariance are fundamental principles.

Based on this paradigm, several simulators have been developed, among which the Graph Network Simulator (GNS) and MeshGraphNet stand out as the most promising. While differing mainly in the graph construction method, they both excel at simulating complex phenomena such as fluid dynamics, thanks to message passing and graph-to-graph updates to generate subsequent time steps of a simulation~\citep{sanchez2020learning,pfaff2020learning,fortunato2022multiscale}.

Despite this progress, applications of GNN surrogates to structural problems, particularly dynamic ones, remain comparatively limited. Existing work often focuses on static or time-independent PDEs, quasi-static responses, or settings where displacements are large relative to the geometric scale~\citep{chou2024structgnn,gladstone2024mesh-based,zhao2024gnnmechanics,deshpande2022graphunet,herrmann2024deeplearning}. However, a general framework capable of handling dynamic events and micro-scale displacements has not yet been proposed. The latter aspect, namely micro-scale displacements, introduces several numerical challenges: finite-difference velocities computed from absolute nodal positions can suffer from catastrophic cancellation when subtracting nearly equal floating-point numbers, which degrades derivative accuracy and destabilizes long rollouts~\citep{higham2002accuracy}.

To address these gaps, we introduce the Graph Network-based Structural Simulator (GNSS), a GNN framework for surrogate modeling of structural dynamics. The design is general and not restricted to any specific structural class; in this paper, we evaluate it on guided wave dynamics, an application that is both practically relevant - for example, in fields like structural health monitoring - and numerically demanding, due to the need for fine meshes and spatial discretizations to obtain accurate results~\citep{rose1999ultrasonicwaves,cawley2024guidedwaves}. GNSS combines three ingredients tailored to structural simulations: (i) a \emph{local-coordinate} formulation that expresses nodal kinematics in node-fixed frames to stabilize finite-difference velocities at micro-scale displacements; (ii) a \emph{sign-aware} acceleration loss that discourages phase flips and improves long-horizon stability; and (iii) a \emph{wavelength-informed} connectivity radius that aligns the message-passing neighborhood with physically meaningful interaction scales (e.g., a fraction of the bending-wave wavelength), thereby leveraging locality without oversmoothing~\citep{langer2017more}. Together, these choices preserve the benefits of graph locality, while mitigating failure modes observed when applying off-the-shelf GNN simulators to structural dynamics problems involving small displacements.

We benchmarked GNSS on a numerical case study involving wave propagation in a clamped beam, using a dataset generated through finite element simulations. Our model accurately predicts wave propagation during the rollout phase and generalizes across different loading conditions, outperforming traditional GNNs based on absolute nodal positions.

This paper is organized as follows. Section~\ref{sec:methods} describes the implementation of GNSS. Section~\ref{sec:case} introduces the case study used to validate the proposed method and discusses the results. Section~\ref{sec:conclusions} draws out the conclusions of this work.

\section{Methods}\label{sec:methods}
GNSS builds on the foundational concepts of graph theory and the classical GNNs introduced by \citet{scarselli2009graph}, as well as the GN framework proposed by \citet{battaglia2018relational}. A comprehensive treatment of these theoretical foundations is beyond the scope of this paper; instead, the focus is on adapting and applying the GN framework to structural dynamics. Accordingly, this section first introduces basic graph definitions, then outlines the GN paradigm, and finally presents the specific GN-based architecture developed for structural dynamic analysis.

\subsection{Graph Networks}
A graph \(\mathcal G\) consists of a set of nodes (or vertices) and a set of edges that define connections between nodes. Formally, a graph can be represented using an adjacency matrix \(\mat A \in \mathbb{R}^{N \times N}\), where \(N\) is the number of nodes, and \(A_{ij} \neq 0\) indicates an edge between nodes \(i\) and \(j\). The degree of a node is defined as the number of edges incident to that node. Beyond this topological knowledge, graphs can include additional information in the form of node features \(\mat V \in \mathbb{R}^{N \times d}\) and edge features \(\mat E \in \mathbb{R}^{|\mat E| \times k}\), where \(d\) and \(k\) are the feature dimensions (i.e., the length of the vector) of nodes and edges, respectively~\citep{gong2019exploiting,yang2020nenn,chen2021edgefeatured}. The node feature vector associated with node \(n\), denoted by \( \vec x_n \in \mathbb{R}^d\), represents some measurable properties of the node. Similarly, the edge feature vectors encode the relationship between connected nodes. Given the feature vectors, the graph can be defined as \(G= (\mat V, \mat E)\)

Early neural architectures for graphs, such as the GNN model proposed in \citet{scarselli2009graph}'s work, extended neural computation to structured graph domains using an iterative, equilibrium-based message-passing scheme. Each node updates its state by aggregating information from its neighbors until convergence. These models have proven effective for representing physical systems with complex, irregular topologies, where traditional architectures like Convolutional Neural Networks (CNNs), designed for structured grid-like data, are inadequate. GNNs generalize the concept of convolution to graph domains, making them well-suited for applications in fluid dynamics, structural mechanics, and materials science~\citep{battaglia2018relational,pfaff2020learning,wong2022graph,sanchez2020learning,choi2024graph,li2023machine,zhao2023intelligent,gulakala2023graph,shivaditya2022graph,bronstein2017geometric,wu2020comprehensive}.

A key strength of GNNs in modeling physical systems lies in their inductive biases. These include the representation of local interactions through message passing, the sharing of weights across the graph, and invariance to node permutations. Such biases naturally reflect physical principles such as locality, translational symmetry, and conservation laws~\citep{battaglia2018relational}.

The GN framework~\citep{battaglia2018relational} generalizes and unifies multiple GNN-based architectures such as message-passing neural networks (MPNNs), interaction networks, and relation networks. Unlike classical GNNs, GNs explicitly decompose the graph update process into three modular functions: edge update, node update, and global update. This modular graph-to-graph transformation enables flexible and expressive modeling capabilities, which are particularly beneficial in physical simulation tasks~\citep{sanchez2020learning}.

Given as input a system's state described by a graph \(\mathcal S = (\mat X, \mat R)\), a GN produces an updated graph \(\mathcal S' = (\mat X', \mat R')\) of identical topology representing the updated state of the system. \(\mat X\) and \(\mat R\) describe the physical properties and relationships of the system, respectively. The architecture typically follows an \textit{encode-process-decode} paradigm~\citep{battaglia2018relational}, where raw inputs are encoded into latent node and edge embeddings, iteratively processed through message passing, and decoded into physically meaningful outputs.

\paragraph{Encoding} The encoding phase is described in Equation~\ref{eq:gnn_encoding}. During this step, the physical properties and relationship describing the state are embedded into a higher-dimensional continuous space:
\begin{equation}\label{eq:gnn_encoding}
    \begin{dcases}
    \mat X \in \mathbb R^{N\times d} \\ 
    \mat R  \in \mathbb R^{N\times k}
    \end{dcases}
    \xrightarrow{} 
    \mat V, \, \mat E \in \mathbb{R}^{N\times h}
\end{equation}
Here, \(N\) is the number of nodes, \(d\) is the number of properties for each node, \(k\) is the number of relationships for each edge, and the \(h\)-dimensional space is the \emph{latent space}. This initial transformation not only homogenizes the feature space, but also allows the network to capture local patterns that are crucial for the learning task~\citep{scarselli2009graph, gilmer2017neuralmessagepassing}.

\paragraph{Message Passing} Message passing is the phase during which information is processed and exchanged among the different nodes of the system. It consists of updating the latent features of the graph through three steps:
\begin{enumerate}
    \item \textbf{Message Construction} (Equation~\ref{eq:gnn_construction}): The updated edge features (also called \emph{message}) are obtained by combining the old information of the edge and the two nodes connected to it: 
    \begin{equation}\label{eq:gnn_construction}
        \vec e^\prime_{ij} = \phi \left(\vec v_i,\vec v_j,\vec e_{ij}\right)
    \end{equation}
    \item \textbf{Message Aggregation} (Equation~\ref{eq:gnn_aggregation}): For each node, an aggregated message is computed; this is obtained by combining all the messages related to node \(i\): 
    \begin{equation}\label{eq:gnn_aggregation}
        \vec{\bar v}_i=\psi(\vec e^\prime_{ij} \,,\;\forall j \in \mathcal{N}(i))
    \end{equation} 
    Here, \(\mathcal{N}(i)\) represents the set of nodes linked to node \(i\).
    \item \textbf{Node Update} (Equation~\ref{eq:gnn_update}): During this final step, the old node feature vector is combined with the aggregated message to obtain the updated node features:
    \begin{equation}\label{eq:gnn_update}
        \vec v^\prime_i = \gamma\left( \vec v_i, \vec{\bar v}_i \right)
    \end{equation}
\end{enumerate}
Equation~\ref{eq:gnn_message_passing} describes the entire message passing pipeline concisely:
\begin{equation}\label{eq:gnn_message_passing}
        \vec v^\prime_i = \gamma\left( \vec v_i, \psi_{j \in \mathcal{N}(i)} \left( \vec \phi \left(\vec v_i,\vec v_j,\vec e_{ij}\right)\right)\right)
\end{equation}
In Equations~\ref{eq:gnn_construction}--\ref{eq:gnn_message_passing}, the symbols \(\phi,\,\psi,\,\gamma\) represent arbitrary operands which are part of the architectural choice when designing a GN. 

\begin{figure}[htb!]
    \centering
    \includegraphics[width=.9\linewidth]{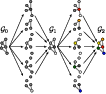}
    \caption[Visual representation of the influence of two message passing steps on information spreading]{Visual representation of information propagation over two message-passing steps. In the second step, colored nodes receive messages from the same neighbors as in the first step, since the topology is fixed. However, these neighbors now carry the information acquired from their own neighbors in the previous step, illustrated using a grayscale colormap.
}
    \label{fig:message_passing_steps}
\end{figure}

The whole message passing procedure is typically repeated \(M>1\) times. This hyperparameter defines the degree of information propagation through the network, and thus the depth of the GN, as visually represented in Figure~\ref{fig:message_passing_steps}. Information spread is usually represented through a computation graph, which describes how the embedding of each node is iteratively updated through message passing. Each layer of the computation graph carries out a message-passing operation; by applying this process recursively across layers, the network is able to represent both local interactions and more complex, higher-order relationships within the graph~\citep{gilmer2017neuralmessagepassing,battaglia2018relational}.

\paragraph{Decoding} During the decoding phase, the information stored in the latent representation is converted to task-specific predictions, such as node-level, edge-level, or graph-level outputs~\citep{battaglia2018relational}. For example, referring to the case of GNS and MeshGraphNet, the decoder performs a single node-level task by transforming the updated node features into a physical quantity at each node, the acceleration.

\subsection{GNSS}
\begin{figure*}[htb!]
    \centering
    \includegraphics[width=\textwidth]{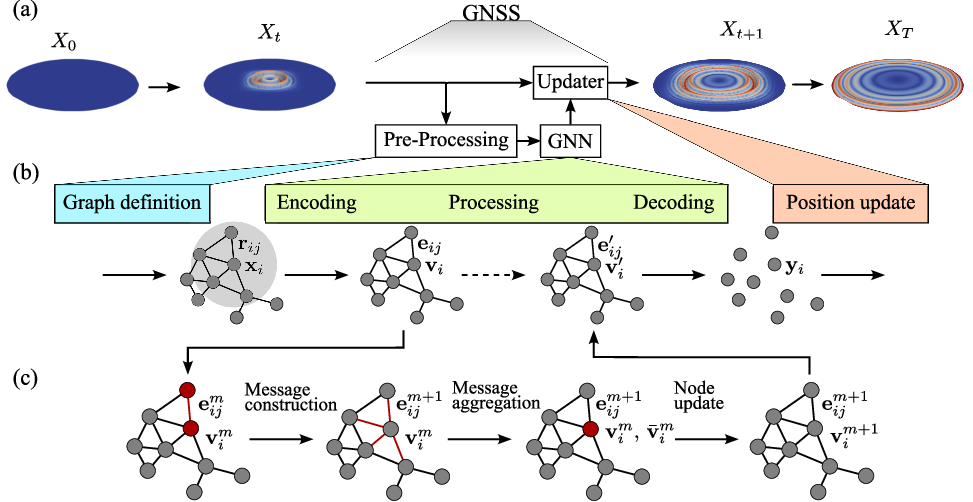}
    \caption{Our GNSS framework, adapted from \citet{sanchez2020learning}'s work. a) Rollout: an initial configuration $X_0$ is provided, then the GNSS is iteratively applied $T$ times to predict the trajectory from time 0 to time T. GNSS includes a graph building procedure, a GNN, and a fixed updater. b) Encode-process-decode procedure. c) Message passing operation at message passing step \(m\) with \(m=1,\dots, M\).}
    \label{fig:framework}
\end{figure*}

In Figure \ref{fig:framework}, our GNSS is presented. This framework is a modified version of the GNS developed by \citet{sanchez2020learning}, specifically adapted for structural simulations. Figure \ref{fig:framework}(a) shows the \textit{Rollout}, namely the procedure used to simulate the trajectory of the system: an initial configuration $X_0$ is provided, then the GNSS is iteratively applied $\frac{T}{\Delta t_{ph}}$ times to predict the trajectory from physical time 0 to physical time \(T\). \(\Delta t_{\mathrm{ph}}\) is the fixed timestep size adopted in the numerical simulations used to generate all trajectories in the dataset. Figure \ref{fig:framework}(b) expands the operation involved at each time step \(t\), during which the GNSS performs the whole encode-process-decode paradigm to retrieve the updated nodal information (accelerations). Lastly, Figure \ref{fig:framework}(c) details the processing phase, i.e., message passing, showing the procedure for a single passing step; as introduced above, message passing is recursively repeated \(M\) times.

In what follows, we detail the GNSS operations shown in Figure~\ref{fig:framework}(b)--(c).
The pipeline — from graph construction to position update — is described for a single
simulation time step \(t\); accordingly, all intermediate quantities are evaluated at \(t\),
and the output corresponds to the updated state at \(t+1\). Because a single GNSS pass
performs multiple rounds of message passing, we denote the \(m\)-th round by the
superscript \(^{(m)}\). To avoid clutter, we omit the time index; unless otherwise stated,
all expressions are understood at time \(t\).

\paragraph{Pre-processing: graph definition}
The pre-processing phase is responsible for representing the information stored in the initial configuration of the system into a graph structure ready to be processed by the GN.

The representation of the system naturally follows from the choice of a graph-based model. The continuous structure is discretized into a set of \emph{points}, each capturing the local physical properties of its neighborhood, while the interactions among these points are expressed as \emph{relationships}. Throughout the following description, we will denote a point in the discretized structure as \(x\), and its associated relationship as \(r_{ij}\), representing the underlying physical properties. In contrast, the terms \emph{node} \(v_i\) and \emph{edge} \(e_{ij}\) will refer to the corresponding latent features, i.e., the latent graph representation used within the model.

\begin{figure}[H]
    \centering
    \includegraphics[width = .4\columnwidth]{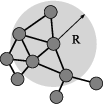}
    \caption{Graph defined through connectivity radius.}
    \label{fig:connectivity}
\end{figure}

Regarding the relationship between those points, the primary idea behind the graph construction proposed by \citet{sanchez2020learning} is that, typically, entities in physical systems are influenced by their neighbors. This natural behavior is reflected in the graph structure by establishing edges only between nodes that are sufficiently close. In practice, this is achieved by defining a connectivity radius that sets the maximum allowable distance for two nodes to be connected, as shown in Figure \ref{fig:connectivity}.

Once the graph's topology is defined, each node \(i\) of the graph is associated with a feature vector storing physical information. Following the implementation proposed by \citet{sanchez2020learning}, the point feature vector stores information about the velocity context and the type of particle related to the node. As demonstrated by \citet{sanchez2020learning}, including not only the current velocity of a point, but also the velocities from previous timesteps, greatly improves the performance of the model. This information is stored in a tensor as shown in Equation~\ref{eq:velocity}:
\begin{equation}\label{eq:velocity}
     \prescript{(t)}{}{\dot{\ten{P}}}_i = \left[\prescript{(t-n)}{}{\dot{\vec p}}_i ,\dots, \prescript{(t)}{}{\dot{\vec p}}_i \right]  \qquad \prescript{(t)}{}{\dot{\ten{P}}}_i \in \mathbb R^{d\times n}
\end{equation}
Here, \(t\) is the current timestep, and \(\prescript{(t)}{}{\dot{\vec p}}_i \) is the velocity of point \(i\) at time \(t\), \(d\) is the dimension of the space, and \(n\) is the number of previous velocities considered (in addition to the current one). This is a hyperparameter, and is usually set to \(n=4\) as suggested by \citet{choi2024graph}. The velocities are calculated via finite differences as shown in Equation~\ref{eq:velocity_finite_diff}:
\begin{equation}\label{eq:velocity_finite_diff}
    \prescript{(t)}{}{\dot{\vec p}}_i  = \frac{\prescript{(t)}{}{{\vec p}}_i  - \prescript{(t-1)}{}{{\vec p}}_i }{\Delta t}
\end{equation}
where \(\prescript{(t)}{}{{\vec p}}_i\) is the position of point \(i\) at time \(t\) and \(\Delta t\) is the timestep size. The timestep size used in the algorithm is set to 1, as non-dimensional time quantities are preferred during prediction. This unit timestep size corresponds to a physical timestep \(\Delta t_{ph}\) implicitly defined by the one used in the training dataset. Consequently, the timestep size adopted in the numerical simulations must remain fixed and identical across all trajectories in the dataset.

The framework is designed to account for different types of particles. This information is required by the model to correctly handle entities playing different roles in the simulation, and is stored in a vector \(\vec f\). This vector has 16 components and is created through an embedding layer. 

In our model, we exploit this structure to inform the prediction mechanism about the constraint of the structural simulation. Thanks to this choice, we can simply assign arbitrary IDs to different types of constraints, and the model will optimize the weights of the embedding layer (i.e., the 16 components of \(\vec f\)) to best represent the different entities according to their behavior. For example, we can assign an ID to the free portion (nodes) of the structure, another ID to nodes where a prescribed motion is applied, and another ID to the region of application of BCs. This choice allows us to simplify the feature vector proposed by \citet{choi2024graph} by eliminating the information about the distance from the boundaries that was given for each node. Additionally, this allows us to model different types of boundaries by assigning different IDs to them (clamp, pin, etc.).

All the properties listed above are included in a point state vector, as shown in Equation~\ref{eq:physical_node_features}:
\begin{equation}\label{eq:physical_node_features}
    \vec x_i^t = \left[\dot{\vec p}_i^{\le t}, \vec f\right]
\end{equation}
Here, \(\dot{\vec p}_i^{\le t}\) is the flattened vector of velocities history. For a representative 2D case, considering the current velocity and 4 additional velocities and the previous timesteps, the point feature vector is composed of 26 elements: 5 components for velocity along direction \(x\), 5 for velocity along direction \(y\), and 16 for the type embedding.

Moreover, the interaction between connected points \(i\) and \(j\) is represented by their distance in the physical space and their displacement in the current timestep, both normalized by the connectivity radius $R$, as shown in Equation~\ref{eq:physical_edge_features}:
\begin{equation}\label{eq:physical_edge_features}
    \vec r_{i,j}^t = \left[\left(\vec p_i^t-\vec p_j^t\right),\left\lVert \vec p_i^t-\vec p_j^t\right\lVert \right]    \frac{1}{R}
\end{equation}
The distance provides information about the level of interaction between the two points, while the displacement gives a direction to the relationship.

Typically, GNs represent displacements in an absolute coordinate system shared by the entire physical system \citep{sanchez2020learning, choi2024graph}. While this approach is well-suited to domains commonly simulated with GNs, such as granular flow and fluid dynamics, it can be problematic in structural mechanics. In structural simulations, load-induced displacements are often several orders of magnitude smaller than the characteristic dimensions of the system. This scale mismatch can cause numerical issues when derivatives (e.g., velocities) are computed via finite differences. In particular, subtracting two large, nearly equal floating-point numbers (absolute positions at successive timesteps) may lead to a loss of significant digits, a phenomenon known as \emph{catastrophic cancellation} \citep{higham2002accuracy}. The resulting numerical noise reduces the accuracy of computed derivatives, and the problem becomes critical when rounding errors are comparable to, or larger than, the displacements being resolved. This limitation is well recognized in precision-sensitive fields such as structural dynamics and geomechanics, where floating-point inaccuracies can distort small-strain calculations \citep{belytschko2014nonlinear}.

To overcome this limitation, GNSS introduces a novel representation of nodal positions in local coordinate systems that are fixed in both space and time. For each node, the origin is set at its initial position (i.e., its location at time zero), effectively centering all subsequent displacements around zero. By eliminating large absolute position values, this representation mitigates catastrophic cancellation and ensures stable, accurate derivative computations via finite differences. This formulation constitutes a key feature of GNSS, enabling reliable surrogate modeling of structural dynamics where traditional GN formulations fail.

\paragraph{Encoder}
The encoder is responsible for transforming the input graph \(\mathcal{G}\) into a latent graph \(\mathcal{G}_0\), in which the old physical feature vectors are embedded into a latent space. In our implementation, following \citet{sanchez2020learning}'s work, this operation is performed by two multilayer perceptrons (MLPs), \(\epsilon^v_{\Theta^v}\) and \(\epsilon^e_{\Theta^e}\), which embed the feature vectors associated with nodes and edges, respectively. The two MLPs share the same architecture: an input layer with dimensions \(N_x\) for nodes and \(N_r\) for edges, two hidden layers of 64 units each, and an output layer of size 64. The sets of trainable parameters are identified by \(\Theta^v\) and \(\Theta^e\), and are optimized during training to learn an effective way to embed physical properties into the latent space. This operation can be represented as shown in Equation~\ref{eq:encoding}:
\begin{equation}\label{eq:encoding}
    \vec v_i^t = \epsilon_\Theta^v\left(\vec x_i^t\right), \quad
    \vec e_{ij}^t = \epsilon_\Theta^e\left(\vec r_{ij}^t\right)
\end{equation}
Here, \(\vec v_i^t\) and \(\vec e_{ij}^t\) represent the node and edge feature vectors in the latent space.

Figure \ref{eq:encoding} shows the encoding procedure for a node and an edge of a sample graph. The result of these operations is a latent graph \(\mathcal G^0= \left(\mat V, \mat E\right)\); the superscript \(^0\) indicates that this is the initial state representation, prior to performing any message-passing step.
\begin{figure}[htb!]
    \centering
    \includegraphics[width = \linewidth]{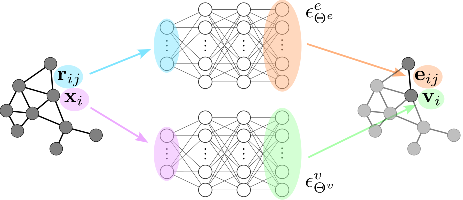}
    \caption{Representation of the encoder block: two MLPs, \(\epsilon^v_{\Theta^v}\) and \(\epsilon^e_{\Theta^e}\), embed the node and edge feature vectors, respectively, into the latent space.
}
    \label{fig:encoding}
\end{figure}

\paragraph{Processor}
The processor performs the message passing operations, a simplified representation of which is shown in Figure~\ref{fig:framework}(c).

\begin{figure}[htb!]
    \centering
    \includegraphics[width=\linewidth]{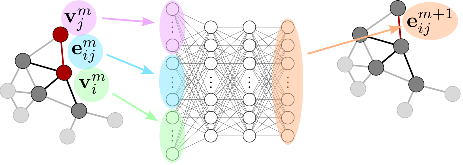}
    \caption{Message construction: an MLP merges the information from each link (2 nodes and 1 edge) into an update edge feature.}
    \label{fig:message_construction}
\end{figure}

The processor takes as input the initial graph \(\mathcal{G}^m\), where the index \(m\) denotes the current message-passing step. This index starts from \(m=0\) (the encoder output) and is incremented at each iteration until a user-defined number \(M\) of steps is reached. Message passing begins with the message construction phase, illustrated in Figure~\ref{fig:message_construction} and described by Equation~\ref{eq:gns_construction}. In this phase, each edge feature vector is updated by combining its current value with the feature vectors of the two nodes it connects, all evaluated at the current iteration. In our model, this combination is implemented by an MLP with two hidden layers of 64 units each. The input layer has size \(3n\), since the MLP input is the concatenation of the three feature vectors involved in constructing the message for each edge. The output layer has size \(n\), corresponding to the latent space dimension, in order to preserve consistency in the feature vector dimensions across the graph entities. This operation can be generalized as:
\begin{equation}\label{eq:gns_construction}
    \vec e^{m+1}_{i,j} = \phi^m_{\vec\Theta^\phi_m}(\vec v_i^m,\vec v_j^m,\vec e_{i,j}^m)
\end{equation}
Here, \(\phi^m\) denotes the MLP used at the \(m\)-th message-passing step, and \(\vec{\Theta}^\phi_m\) is the set of trainable parameters associated with it. The operation is repeated for all edges in the graph.

\begin{figure}[htb!]
    \centering
    \includegraphics[width=\linewidth]{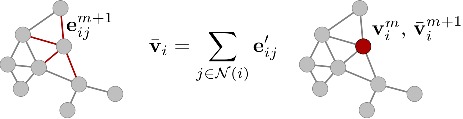}
    \caption{Message aggregation: all the (updated) information coming from neighboring nodes is gathered.}
    \label{fig:message_aggregation}
\end{figure}

The output of the message construction phase is then used for message aggregation, as illustrated in Figure~\ref{fig:message_aggregation} and described by Equation~\ref{eq:gns_aggreg}. For each node \(i\) in the graph, an aggregated message is constructed by element-wise summing the feature vectors of the edges connected to it. This is expressed as:
\begin{equation}\label{eq:gns_aggreg}
    \bar{\vec v}_i^{m+1} = \sum_{j\in N(i)} \vec e^{m+1}_{i,j}
\end{equation}
For each node, the aggregated message is computed and stored together with its previous feature representation in the graph.

\begin{figure}[htb!]
    \centering
    \includegraphics[width=\linewidth]{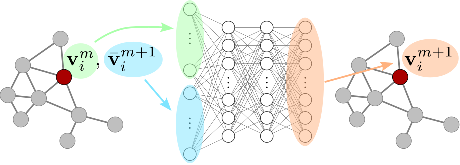}
    \caption{Node features update: the old node feature vector and the aggregated message are processed to obtain the updated node feature vector.}
    \label{fig:node_update}
\end{figure}

Finally, the node features are updated, as illustrated in Figure~\ref{fig:node_update} and by Equation~\ref{eq:gns_update}. This step involves combining the old node feature vector with the aggregated message to obtain the updated node feature vector. In our model, following the structure from \citet{sanchez2020learning}, this operation is performed by an MLP with two hidden layers of 64 units each. The input layer has size \(2n\) (i.e., two times the latent space dimension, since the aggregated message has dimension \(n\)), and the output layer has dimension \(n\). This operation is defined as:
\begin{equation}\label{eq:gns_update}
    \vec v^{m+1}_i = \gamma^m_{\vec \Theta^\gamma_m}(\vec v^m_i, \bar{\vec v}^{m+1}_i)
\end{equation}
Here, \(\gamma^m\) is an operator representing the MLP associated with the current message-passing step, and \(\vec \Theta_\gamma\) represents the set of trainable parameters associated with the MLP. The operation is repeated for all nodes in the graph.

In our model, \(M\) is set to 10, as in \citet{choi2024graph}'s work. Two main aspects cause this hyperparameter to be particularly critical for the model's performance. The first consideration concerns information propagation: a lower \(M\) may not allow information to reach distant parts of the graph, potentially missing important relational data. Conversely, too many message passing rounds can distort relevant signals, leading to less effective representations. Another important consideration is model complexity. Each message passing step uses its own unique MLPs. This means that with every additional step, new parameters are introduced. As a result, the number of parameters grows linearly with the number of message passing steps, and the computational cost increases as each MLP must be executed in sequence.

At the end of all the prescribed message passing steps, the output of the processor block is an updated graph \(\mathcal G'=\mathcal G^M\) containing the update node feature vector \({\vec{v}^\prime_i}\) and the updated edge feature vector \({\vec{e}^\prime_{ij}}\).

\begin{figure}[htb!]
    \centering
    \includegraphics[width=\linewidth]{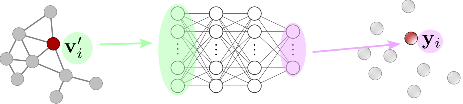}
    \caption{Decoding: each updated node features vector is transformed into the desired physical quantity at the corresponding node in space (acceleration in case of structural simulations).}
    \label{fig:decoding}
\end{figure}

\paragraph{Decoder}
The decoder extracts the dynamics of all points in the physical system, which correspond one-to-one to the graph nodes, from the information stored in the updated graph \(\mathcal{G}^\prime\). In other words, the extracted information represents the output of the procedure and corresponds to the prediction for the next time step, \(t+1\), as shown in Equation~\ref{eq:decoding}:
\begin{equation}\label{eq:decoding}
    \prescript{(t+1)}{}{\vec y_i} = \delta^v_{\vec \Theta}\left( \vec v^\prime_i \right)
\end{equation}
Here, \(\vec v^\prime_i\) represents the updated node features of node \(i\), while \( \prescript{(t+1)}{}{\vec y_i}\) represents the predicted dynamics. For structural simulations, the model is trained to predict the acceleration of all points in the physical system. Accordingly, \(\prescript{(t+1)}{}{\vec y_i}\) is a vector of dimension \(d\),matching the dimension of the physical space. The operator \(\delta^v_{\vec \Theta}\) is the decoder: in our model, this function is an MLP with two hidden layers of 128 units each, with an input layer having the size of the latent space (i.e. \(n\)), and an output layer having size \(d\).
The decoding process is illustrated in Figure~\ref{fig:decoding}.

\paragraph{Updater}
The updater enforces an inertial frame by incorporating the laws of motion through inertial and static priors. The inertial prior assumes that the velocity of each point changes predictably based on its acceleration over a short time interval (Equation~\ref{eq:inertial_prior}), while the static prior assumes that the position of each point evolves continuously with its current velocity (Equation~\ref{eq:static_prior}). These priors are enforced by hardcoding the updater to calculate the new velocity and position via Euler integration:
\begin{align}
    \prescript{(t+1)}{}{\dot{\vec p}_i} &= \prescript{(t)}{}{\dot{\vec p_i}} + \prescript{(t+1)}{}{\vec y_i} \Delta t, \label{eq:inertial_prior} \\
    \prescript{(t+1)}{}{\vec p_i} &= \prescript{(t)}{}{\vec p_i} + \prescript{(t+1)}{}{\dot{\vec p}_i} \Delta t, \label{eq:static_prior}
\end{align}
where \(\prescript{(t+1)}{}{\vec y_i}\) is the predicted acceleration of point \(i\).

\subsubsection{Procedures}
Our framework relies on two core procedures: \emph{training} and \emph{rollout}. The training phase produces a learned model capable of predicting positions, while the rollout phase acts as a solver, generating full trajectories from user-provided input data.

\paragraph{Training}
The training procedure is summarized in the pseudo-code shown in Algorithm~\ref{alg:training}.
\begin{algorithm}[htb]
\small
\caption{Training Procedure}
\label{alg:training}
\SetAlgoLined
\KwIn{Training dataset}
\KwOut{Trained GNSS}
Initialize model\;
\ForEach{training step}{
    Sample a batch of $B$ timesteps\;
    \ForEach{sample \textnormal{in} batch}{
        Calculate velocity from the configuration;
        Add noise\;
        Compute input graph $\mathcal{G} = (\mat X, \mat R)$\;
        Apply the GN block to obtain the update graph $\mathcal G'$\;
        Extract the predicted acceleration $\vec y$ from $\mathcal G'$\;
        Compute true acceleration $\hat {\vec y}$ from ground truth with noise\;
        Compute loss\;
    }
    Backpropagate and update weights\;
}
Save model, i.e. the trained GNSS\;
\end{algorithm}
\begin{figure}[b!]
    \centering
    \includegraphics[width=.9\linewidth]{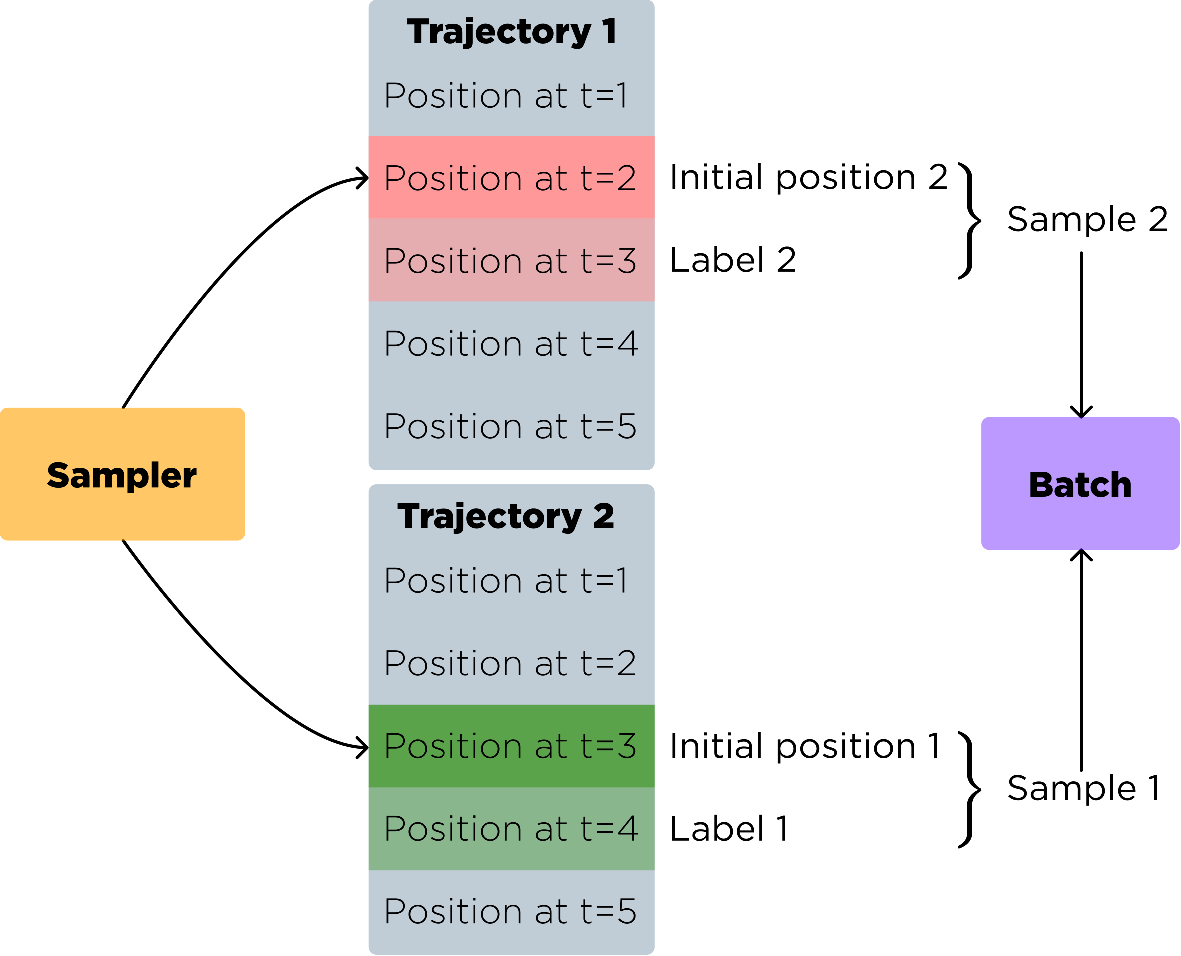}
    \caption{Batching operation for a representative dataset of two trajectories with five timesteps each, and a batch size of two samples. Each sample contains the input (initial) configuration and the label, i.e., the (next) configuration to be predicted.}
    \label{fig:batching}
\end{figure}
As in traditional ML algorithms, training proceeds over successive epochs. At each epoch, the sampler extracts a batch of data from the training set. Each batch consists of \(B\) samples, randomly selected from all timesteps across all trajectories. Figure~\ref{fig:batching} illustrates the batching operation for a representative dataset with two trajectories of five timesteps each. In this example, each batch contains \(B=2\) random samples across all the available data. In the interest of simplicity, each sample consists of the system configuration at timestep \(t\) together with that at timestep \(t+1\). However, in practical applications, information from several previous timesteps is often included to improve temporal context and prediction accuracy. The configuration at \(t\) is used as input to the model, while the acceleration required to evolve the system from configuration \(t\) to configuration \(t+1\) ($\prescript{(t+1)}{}{\vec y}$) serves as the ground truth (label) for the current training step. 

From the input configuration (positions), velocities are extracted via finite differences. To improve robustness against error accumulation and instabilities during rollout, Gaussian noise is added to all computed velocities in the training samples, following \citet{sanchez2020learning}. This regularization strategy encourages the model to generalize beyond the exact training trajectories and mitigates the buildup of prediction errors over time.

After noise injection, GNSS encodes the system configuration at timestep \(t\) from each sample into a graph, predicts the corresponding acceleration ($\vec y$), and compares it with the ground truth (\(\hat {\vec y}\)). Backpropagation is then used to update the weights of all MLPs and the embedding layer in order to minimize a user-defined loss function. To further improve generalization in long rollouts, we propose a novel loss function, the weighted Mean Squared Error (wMSE) shown in Equation~\ref{eq:wMSE}: a modified version of the Mean Squared Error(MSE) that penalizes acceleration predictions with incorrect sign through a scalar weight \(s\), defined as:
\begin{equation}\label{eq:wMSE}
    \varepsilon_i = 
    \begin{cases}
        y_i - \hat{y}_i, & \text{if } y_i \cdot \hat{y}_i \ge 0, \\[6pt]
        s \cdot (y_i - \hat{y}_i), & \text{otherwise}.
    \end{cases}
\end{equation}
Here, the subscript \(_i\) indicates the \(i-th\) node; the final loss is computed as the mean squared value of \(\varepsilon_i\) over all predictions in the batch.

\begin{table*}[b!]
\centering
\small
\begin{tabular}{lll} 
\toprule
\textbf{Module} & \textbf{Property} & \textbf{Value}\\ \midrule
\multirow{3}{*}{\textbf{Section}} 
    & Type & Beam \\
    & Profile & Rectangular \\
    & Dimensions & \(\text{width} = \SI{5}{\mm}\), \(\text{height} = \SI{1}{\mm}\) \\ \midrule
\multirow{3}{*}{\textbf{Material}} 
    & Density & \SI{2900}{\kg\per\cubic\m} \\
    & Young's Modulus & \SI{72}{\giga\pascal} \\
    & Poisson Ratio & 0.3 \\ \midrule
\multirow{2}{*}{\textbf{Load}} 
    & BCs & Encastre at both end nodes \\
    & Excitation & Prescribed motion applied to a interior node \\ \midrule
\multirow{3}{*}{\textbf{Step}} 
    & Time Period & \SI{100}{\micro\s} \\
    & Increment Size & \SI{0.1}{\micro\s} \\
    & Step Type & Dynamic, Explicit \\ \midrule
\multirow{2}{*}{\textbf{Mesh}} 
    & Element size & \SI{0.8}{\mm} \\
    & Element Type & Timoshenko beam elements (B21)\\ 
\bottomrule
\end{tabular}
\caption{Details of the finite element simulation setup used to generate the dataset.}
\label{table:fem}
\end{table*}

\paragraph{Rollout}\label{par:rollout}
After training, predictions are generated in an autoregressive manner: the initial state is encoded as a graph, and GNSS is iteratively applied to predict the full trajectory. The number of iterations $\rm{N_T}$ is defined by the user to cover the entire physical timeframe of interest, with the understanding that the physical timestep size is fixed and identical to that used to generate the training dataset.  The corresponding pseudo-code for the rollout procedure is reported in Algorithm~\ref{alg:rollout}.

\begin{algorithm}[htb!]
\small
\caption{Rollout Procedure}
\label{alg:rollout}
\SetAlgoLined
\KwIn{Initial configuration, trained GNSS, number of timesteps $\rm{N_T}$}
\KwOut{Predicted trajectory}
\For{$i = 0$ \KwTo $N_T-1$}{
    Extract velocity from the configuration;
    Compute graph $\mathcal{G} = (\mat X, \mat R)$\;
    Predict accelerations\;
    Update positions using Equations \ref{eq:inertial_prior}-\ref{eq:static_prior}\;
    Store updated configuration\;
}
Return full predicted rollout\;
\end{algorithm}

\section{Case Study and Results}\label{sec:case}

To evaluate the capabilities of GNSS, we generated a validated numerical dataset using Abaqus for a fully clamped beam of length 320\,mm. The beam was excited through a transverse prescribed motion applied to an interior node. The displacement–time history at the input node consists of a single sine-wave cycle at \(50\,\mathrm{kHz}\), modulated by a Hanning window. Table~\ref{table:fem} summarizes the simulation parameters used to create the dataset.

\begin{figure*}[htb!]
    \centering
    \includegraphics[width=0.8\textwidth]{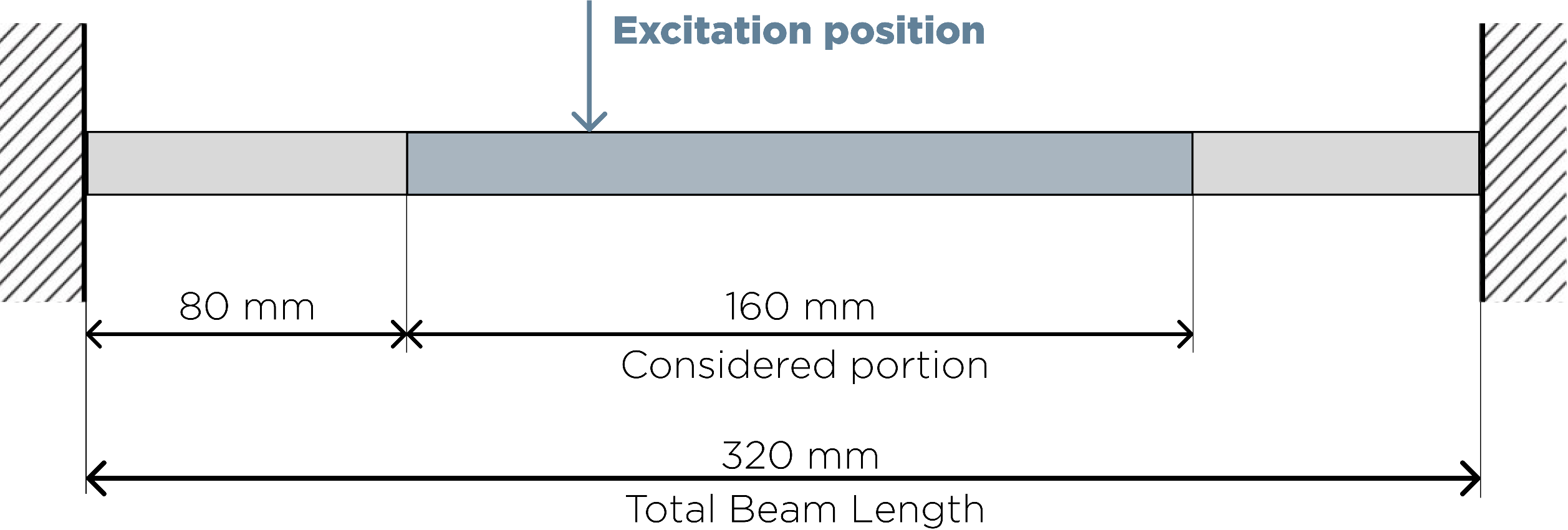}
    \caption{Schematic of the beam configuration. The highlighted section was used to construct the dataset, while boundary regions were excluded to avoid wave reflections.}
    \label{fig:beam_scheme}
\end{figure*}

Six distinct trajectories were generated by varying the location of the prescribed motion, as illustrated in Figure~\ref{fig:beam_scheme}. Only the interior nodes highlighted in the figure were included, while regions near the boundaries were excluded to avoid wave reflections that could interfere with the signal. This choice allows the analysis to focus on pure wave propagation and simplifies the initial case study.Four trajectories were used for training, one for validation, and one was reserved exclusively for testing.

\subsection{Hyperparameter Analysis}

\begin{figure*}[b!]
    \centering
    \includegraphics[width=\textwidth]{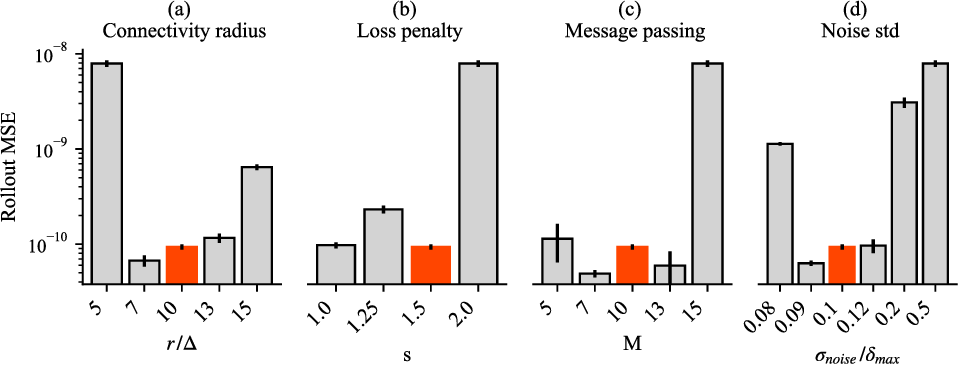}
    \caption{Histograms of the average rollout MSE loss across all trajectories for different hyperparameter settings. Error bars indicate the standard deviation. Red bars correspond to the baseline hyperparameters shared across the sensitivity analysis experiments.}
    \label{fig:sensitivity}
\end{figure*}
A sensitivity analysis was conducted to obtain a coarse tuning of the model hyperparameters.The results are shown in Figure~\ref{fig:sensitivity}.

Red bars correspond to the baseline hyperparameters, which are kept fixed across all experiments except for the one under analysis. Each gray bar therefore represents the outcome of a training and rollout in which a single hyperparameter is varied from its baseline value, while all others remain at their baseline configuration. For example, in panel (a), the bar associated with \(r/\Delta = 7\) corresponds to a model trained with baseline values for all other hyperparameters, but with the connectivity radius set to 7. 

The bar height indicates the rollout MSE loss, defined as the mean Euclidean distance between predicted and ground-truth nodal positions, averaged over all timesteps of the rollout and across all trajectories in the dataset. Error bars denote the corresponding standard deviations.

Panel (a) reports the performance as a function of the connectivity radius. Prior studies suggest that an effective trade-off for modeling fluid and granular systems is to maintain approximately 15–20 edges per node~\citep{sanchez2020learning, choi2024graph}. In structural problems, however, a physically meaningful scale can be identified, directly tied to the choice of the connectivity radius. In particular, the wavelength \(\lambda\) of bending waves in beams can be estimated using the Euler--Bernoulli beam dispersion relation:
\begin{equation}
\lambda = 2\pi \left( \frac{EI}{\rho A (2\pi f)^2} \right)^{1/4},
\end{equation}
where \(E\) is the Young’s modulus, \(I\) the second moment of area, \(\rho\) the material density, \(A\) the cross-sectional area, and \(f\) the wave frequency. For the physical properties of the beam considered here, this yields a wavelength of approximately \(\lambda \approx 13.4\,\mathrm{mm}\). Given our mesh resolution of \(0.8\,\mathrm{mm}\) per node, each wavelength spans approximately 16–17 nodes, and half of this number is sufficient to capture a representative portion of the wave. This is confirmed by the sensitivity analysis, which shows the best performance for a connectivity radius including 7–10 neighboring nodes. Such a choice ensures that local interactions are sufficiently informed by the underlying physics, a key factor for accurately modeling wave propagation dynamics.

Panel (b) shows the performance with respect to the penalty term \(s\) introduced in the proposed loss function. The rollout MSE remains stable for small values of \(s\), with the best performance obtained for \(s = 1.5\).

Panel (c) reports the rollout MSE as a function of the connectivity radius. Although the average MSE is lower with 13 message passing steps than with 10, the computational cost and the variability of the error across different trajectories are significantly higher in the former case. Moreover, while 7 message passing steps yield the best performance, the choice of 10 message passing steps --- motivated by previous studies~\citep{sanchez2020learning, choi2024graph} --- is confirmed as a reliable default setting, which we adopt as the standard in this work.

Finally, panel (d) illustrates the effect of the Gaussian noise standard deviation. The most effective setting is found to be around 9-10\% of the maximum displacement amplitude. Larger noise levels prevent the model from converging, whereas smaller levels do not provide sufficient robustness for long rollouts.

The sensitivity analysis results show that the identified baseline values provide sufficiently accurate performance in terms of rollout MSE. Consequently, these values were adopted as the default configuration and used to train the GNSS model in the subsequent experiments.

\subsection{Training}

\begin{figure}[htb!]
\centering
    \includegraphics[width=\linewidth]{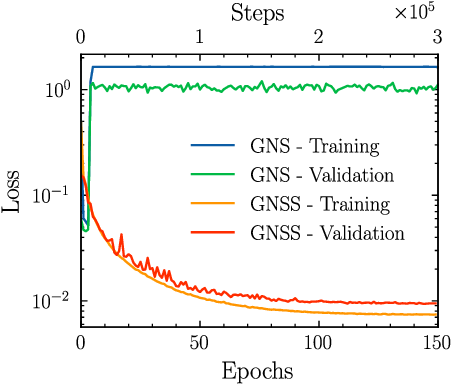}
\caption{Comparison of the training loss between GNSS and the standard GNS.}
\label{fig:training_comp}
\end{figure}

Two models, namely, GNSS and a traditional GNS, were trained using the baseline hyperparameters identified through the sensitivity analysis. The training performance is reported in Figure~\ref{fig:training_comp}. GNSS exhibited a promising decreasing trend during the early epochs, followed by a plateau in the later stages of training. The validation loss confirmed that no overfitting occurred. In contrast, the GNS loss decreased during the first few epochs but subsequently increased, converging to a value higher than the initial one. This behavior suggests that GNS is not effectively learning to capture the underlying physical phenomena, despite the absence of overfitting.

We attribute the difference in performance primarily to the nature of the input information: while GNS operates on variables expressed in absolute coordinates, GNSS encodes the same information in relative coordinates.
\subsection{Rollout}
\begin{figure*}[htb!]
\centering
    \includegraphics[width=\textwidth]{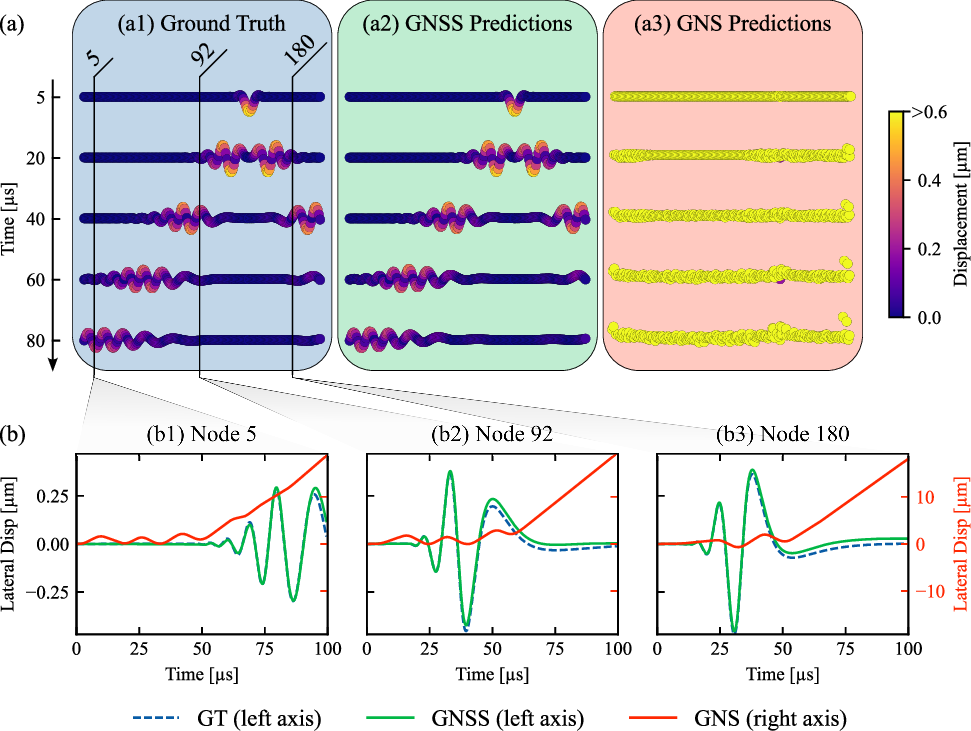}
\caption{Rollout results comparison between GNSS and standard GNS on the test trajectory. (a) Snapshots of the rollouts at five different timesteps for ground truth (a1), GNSS (a2), and GNS (a3), with colorbar indicating the displacement magnitude. (b) Displacement-time histories in the transverse direction for three selected nodes: near the left end (b1), near the midpoint (b2), and near the right end (a3).}
\label{fig:results}
\end{figure*}
The trained models were then evaluated on unseen trajectories, characterized by the same structure and excitation type but applied at different excitation points. The rollout results for both models are presented in Figure~\ref{fig:results}.

Panel (a) presents five rollout snapshots at representative timesteps for the ground truth (a1), GNSS (a2), and GNS (a3). Transverse displacements are on the order of \(\mu\mathrm{m}\), but are visually magnified in the figure for clarity. The colorbar indicates the displacement magnitude, with its range defined by the minimum and maximum values of the ground truth in Figure~\ref{fig:results}(a1).

Qualitatively, the GNSS predictions closely follow the ground truth across all timesteps. In contrast, the GNS model fails to reproduce the physical response. For visualization purposes, the GNS results were scaled by a factor of 0.03 to enable a qualitative comparison of the deformation shapes. The colorbar further indicates that most predicted displacements lie far outside the physical range, underscoring the inability of the GNS model to produce physically consistent results.

Panel (b) provides a quantitative comparison of the displacement–time histories for three representative nodes: node 5 near the left end (b1), node 92 near the midpoint (b2), and node 180 near the right end (b3). Each plot reports the ground truth together with the GNSS and the GNS prediction. A separate vertical axis is used for the GNS curves to enable meaningful visual comparison. The GNSS model shows high accuracy in predicting the trajectory of all considered nodes. In contrast, GNS not only fails to capture the correct waveform, but also diverges to displacement values several orders of magnitude larger than the ground truth.

\begin{figure*}[htb!]
\centering
    \includegraphics[width=\textwidth]{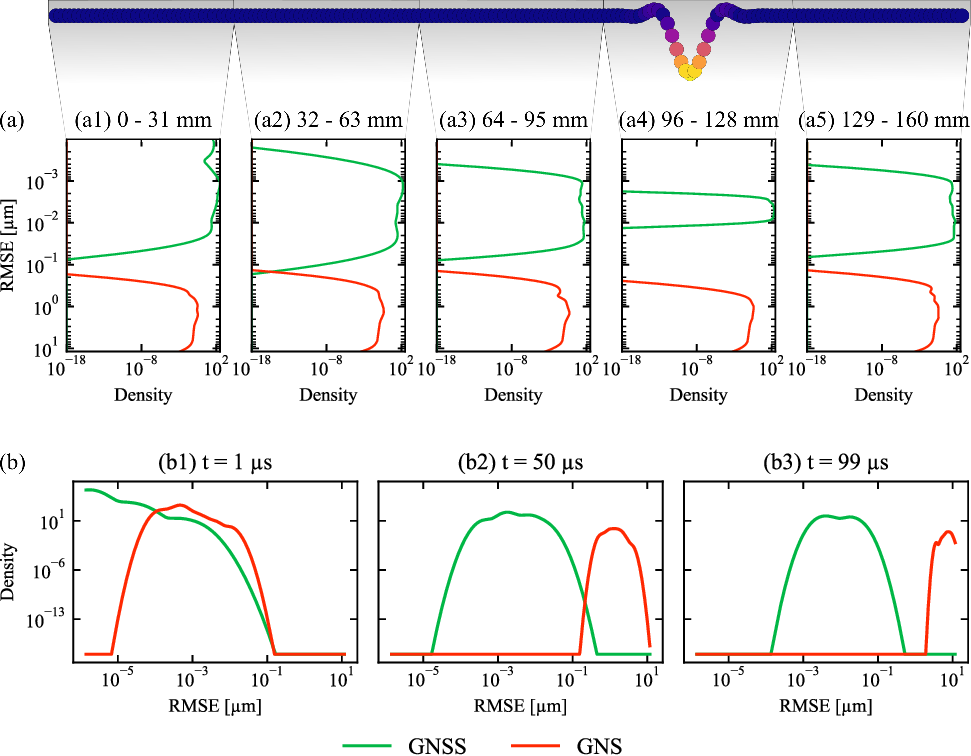}
\caption{RMSE distributions for GNSS (green) and GNS (red). 
(a) Spatial analysis: the beam is partitioned into five equal segments; for each segment, we plot the kernel density of the RMSE averaged over \(30\le t\le 100\,\mu\text{s}\). Sub-panel (a4) contains the segment where the prescribed displacement is applied; the actuated node is excluded from the computation. (b) Temporal accumulation: RMSE distributions aggregated from \(t=0\) to \(t=t_i\) for \(t_i=\{1,50,99\}\,\mu\text{s}\).}
\label{fig:kde}
\end{figure*}

Figure~\ref{fig:kde} summarizes the spatial and temporal distributions of the root-mean-squared error (RMSE) between the predicted and reference values. Panel (a) divides the beam into five equal segments and, for each segment, shows the kernel density of the RMSE averaged over \(t \in [30, 100]\,\mu\text{s}\) to suppress initial-transient bias. Sub-panel (a4) corresponds to the segment containing the node where the prescribed displacement is applied; this actuated node is excluded from the statistics. The error peaks in the input segment and decreases with distance from it, which is consistent with the presence of stronger local gradients and earlier motion near the excitation. Regions farther from the input remain quiescent for longer periods and therefore accumulate less error during the rollout. Panel (b) characterizes error growth over time by plotting the RMSE distributions aggregated from \(t = 0\) up to \(t = t_i\) for \(t_i \in \{1, 50, 99\}\,\mu\text{s}\). Both models accumulate error as the rollout progresses; however, GNSS exhibits only a modest shift toward higher RMSE values and remains concentrated at much lower error levels, whereas GNS shifts markedly and develops heavy tails at larger RMSE. Across both panels, GNSS consistently maintains substantially lower errors than the standard GNS.

The results demonstrate that the standard approach based on absolute positions is inadequate for accurately modeling displacement, velocity, or acceleration fields when displacements are several orders of magnitude smaller than the characteristic structural dimensions. By contrast, GNSS, through its relative coordinate formulation, successfully overcomes this limitation and achieves physically consistent predictions.

\subsection{Runtime Performance}
\label{subsec:runtime_analysis}
On our dataset, GNSS attains an average speed-up of about \(5\times\) over FEM. According to the graph construction method adopted in this work, where two nodes are connected if their distance is less than or equal to a maximum allowable distance (the connectivity radius \(R\)), the expected node degree can be computed as shown in Equation \ref{eq:expected_degree} \citep{penrose2003random}:
\begin{equation}\label{eq:expected_degree}
    \mathbb{E}[\deg]\approx \lambda\,\kappa_d\,R^d
\end{equation}
Here, \(\kappa_d\) is the volume of a unit ball in the \(d\)-dimensional space (\(\kappa_1=2,\, \kappa_2=\pi,\, \kappa_3=\frac{4}{3}\pi\)). Given fixed node intensity \(\lambda\) (analogous to mesh density) and fixed connectivity radius \(R\), the expected value of the degree of the node is bounded, hence the number of edges grows linearly with the total number of nodes, \(E=\Theta(N)\). A message-passing layer with sparse edgewise aggregation therefore costs \(O(E+N)=O(N)\); stacking a fixed number of layers keeps one GNSS rollout step near-linear in \(N\) (up to feature-width constants). 

In contrast, explicit solvers for structural dynamics problems perform time integration using an explicit central-difference scheme with a lumped mass matrix, and the per-increment computational cost scales approximately linearly with the number of elements or nodes~\citep{abaqus_gsx_v66}. However, the stable time increment in explicit integration is limited by a Courant-Friedrichs-Lewy (CFL) condition, which ensures that numerical information does not propagate faster than the physical wave speed within an element. This stability bound scales with the smallest element size, as shown in Equation \ref{eq:courant}:
\begin{equation}
\Delta t_{\mathrm{stable}} \propto \frac{h_{\min}}{c},
\label{eq:courant}
\end{equation}
where $h_{\min}$ is the minimum characteristic element length and $c$ is the relevant wave propagation speed in the material~\citep{abaqus_gsx_v66}. As indicated by Equation ~\ref{eq:courant}, refining a fixed spatial domain (i.e., decreasing element size $h$) simultaneously increases the number of nodes $N \sim h^{-d}$ and decreases the stable time increment $\Delta t_{\mathrm{stable}} \sim h$. Over a fixed physical duration $\tau$, the number of required time steps then scales as $1 / \Delta t_{\mathrm{stable}} \sim h^{-1}$. Combining the per-step computational cost $O(N)$ with $O(1 / \Delta t_{\mathrm{stable}})$ steps yields a total explicit runtime of $O\!\big(N^{1 + 1/d}\big)$, representing superlinear growth imposed by the stability constraint.  

In contrast, GNSS advances the solution through feed-forward message passing at the prescribed time step $\Delta t_{\mathrm{ph}}$ of the training dataset, which remains fixed across trajectories. There is no CFL-type restriction, so the temporal resolution is arbitrarily selected and imposed during training rather than being dictated by numerical stability. The per-step computational cost remains approximately linear in $N$, making the observed $5\times$ speed-up in our simple test case a conservative estimate for higher-resolution problems.

\section{Conclusions}\label{sec:conclusions}
In this work, we have presented GNSS, a graph-network–based surrogate model for time-resolved structural dynamics simulations. The model builds on the GNS framework, originally developed for granular flow and fluid dynamics, and introduces the following key innovations:

\begin{itemize}
  \item \textbf{Relative reference system:} the states of the system are expressed in a relative, node-fixed local coordinate system rather than in the global frame.
  \item \textbf{Novel loss function:} a weighted MSE loss (wMSE) was introduced to penalize acceleration predictions with incorrect sign, enabling robust long-horizon rollouts and accurate spatial field predictions.
  \item \textbf{Physics-based hyperparameter setup:} the sensitivity analysis demonstrated that hyperparameters can be selected based on physical considerations. In the elastodynamic case study, for instance, the connectivity radius was determined from the excitation wavelength. 
\end{itemize}

The case study showed that GNSS successfully predicts wave propagation, a task where the state-of-the-art GNS tailored to granular flow and fluid dynamics fails to provide physically meaningful results.

While the results are promising, further validation is required on additional structural dynamics applications and more complex geometries. Ongoing work includes extending GNSS to three-dimensional elastodynamic problems involving isotropic and anisotropic materials, as well as incorporating information about structural anomalies such as damage. Finally, future efforts will focus on training the framework directly on experimental data, thereby removing the need for numerical simulations and paving the way for structural health monitoring applications.


\end{document}